\documentclass{article} 
\usepackage{times}
\usepackage{wrapfig}
\usepackage{amsthm,amsmath,bbm} 
\usepackage[psamsfonts]{amssymb}
\usepackage{algorithm,algorithmic}
\usepackage[utf8]{inputenc}
\usepackage{graphicx,subfigure}
\usepackage[numbers]{natbib}

\title{Deep Self-Taught Learning for Handwritten Character Recognition}
\author{
Frédéric  Bastien \and
Yoshua  Bengio \and
Arnaud  Bergeron \and
Nicolas  Boulanger-Lewandowski \and
Thomas  Breuel \and
Youssouf  Chherawala \and
Moustapha  Cisse \and 
Myriam  Côté \and 
Dumitru  Erhan \and
Jeremy  Eustache \and
Xavier  Glorot \and 
Xavier  Muller \and
Sylvain  Pannetier Lebeuf \and
Razvan  Pascanu \and 
Salah  Rifai \and 
Francois  Savard \and 
Guillaume  Sicard 
}
\date{June 3, 2010, Technical Report 1353, Dept. IRO, U. Montreal}

\begin{document}

\maketitle

\begin{abstract}
  Recent theoretical and empirical work in statistical machine learning has demonstrated the importance of learning algorithms for deep architectures, i.e., function classes obtained by composing multiple non-linear transformations. Self-taught learning (exploiting unlabeled examples or examples from other distributions) has already been applied to deep learners, but mostly to show the advantage of unlabeled examples. Here we explore the advantage brought by {\em out-of-distribution examples}.  For this purpose we developed a powerful generator of stochastic variations and noise processes for character images, including not only affine transformations but also slant, local elastic deformations, changes in thickness, background images, grey level changes, contrast, occlusion, and various types of noise. The out-of-distribution examples are obtained from these highly distorted images or by including examples of object classes different from those in the target test set.  We show that {\em deep learners benefit more from out-of-distribution examples than a corresponding shallow learner}, at least in the area of handwritten character recognition. In fact, we show that they beat previously published results and reach human-level performance on both handwritten digit classification and 62-class handwritten character recognition.
\end{abstract}

\section{Introduction}

{\bf Deep Learning} has emerged as a promising new area of research in
statistical machine learning (see~\citet{Bengio-2009} for a review).
Learning algorithms for deep architectures are centered on the learning
of useful representations of data, which are better suited to the task at hand,
and are organized in a hierarchy with multiple levels.
This is in part inspired by observations of the mammalian visual cortex, 
which consists of a chain of processing elements, each of which is associated with a
different representation of the raw visual input. In fact,
it was found recently that the features learnt in deep architectures resemble
those observed in the first two of these stages (in areas V1 and V2
of visual cortex)~\citep{HonglakL2008}, and that they become more and
more invariant to factors of variation (such as camera movement) in
higher layers~\citep{Goodfellow2009}.
Learning a hierarchy of features increases the
ease and practicality of developing representations that are at once
tailored to specific tasks, yet are able to borrow statistical strength
from other related tasks (e.g., modeling different kinds of objects). Finally, learning the
feature representation can lead to higher-level (more abstract, more
general) features that are more robust to unanticipated sources of
variance extant in real data.

{\bf Self-taught learning}~\citep{RainaR2007} is a paradigm that combines principles
of semi-supervised and multi-task learning: the learner can exploit examples
that are unlabeled and possibly come from a distribution different from the target
distribution, e.g., from other classes than those of interest. 
It has already been shown that deep learners can clearly take advantage of
unsupervised learning and unlabeled examples~\citep{Bengio-2009,WestonJ2008-small},
but more needs to be done to explore the impact
of {\em out-of-distribution} examples and of the multi-task setting
(one exception is~\citep{CollobertR2008}, which uses a different kind
of learning algorithm). In particular the {\em relative
advantage} of deep learning for these settings has not been evaluated.
The hypothesis discussed in the conclusion is that a deep hierarchy of features
may be better able to provide sharing of statistical strength
between different regions in input space or different tasks.

Whereas a deep architecture can in principle be more powerful than a
shallow one in terms of representation, depth appears to render the
training problem more difficult in terms of optimization and local minima.
It is also only recently that successful algorithms were proposed to
overcome some of these difficulties.  All are based on unsupervised
learning, often in an greedy layer-wise ``unsupervised pre-training''
stage~\citep{Bengio-2009}.  One of these layer initialization techniques,
applied here, is the Denoising
Auto-encoder~(DA)~\citep{VincentPLarochelleH2008-very-small} (see Figure~\ref{fig:da}), 
which
performed similarly or better than previously proposed Restricted Boltzmann
Machines in terms of unsupervised extraction of a hierarchy of features
useful for classification. Each layer is trained to denoise its
input, creating a layer of features that can be used as input for the next layer.  


%
In this paper we ask the following questions:

$\bullet$ 
Do the good results previously obtained with deep architectures on the
MNIST digit images generalize to the setting of a much larger and richer (but similar)
dataset, the NIST special database 19, with 62 classes and around 800k examples?

$\bullet$ 
To what extent does the perturbation of input images (e.g. adding
noise, affine transformations, background images) make the resulting
classifiers better not only on similarly perturbed images but also on
the {\em original clean examples}? We study this question in the
context of the 62-class and 10-class tasks of the NIST special database 19.

$\bullet$ 
Do deep architectures {\em benefit {\bf more} from such out-of-distribution}
examples, i.e. do they benefit more from the self-taught learning~\citep{RainaR2007} framework?
We use highly perturbed examples to generate out-of-distribution examples.

$\bullet$ 
Similarly, does the feature learning step in deep learning algorithms benefit {\bf more}
from training with moderately {\em different classes} (i.e. a multi-task learning scenario) than
a corresponding shallow and purely supervised architecture?
We train on 62 classes and test on 10 (digits) or 26 (upper case or lower case)
to answer this question.

Our experimental results provide positive evidence towards all of these questions,
as well as classifiers that reach human-level performance on 62-class isolated character
recognition and beat previously published results on the NIST dataset (special database 19).
To achieve these results, we introduce in the next section a sophisticated system
for stochastically transforming character images and then explain the methodology,
which is based on training with or without these transformed images and testing on 
clean ones. We measure the relative advantage of out-of-distribution examples
(perturbed or out-of-class)
for a deep learner vs a supervised shallow one.
Code for generating these transformations as well as for the deep learning 
algorithms are made available at {\tt http://hg.assembla.com/ift6266}.
We estimate the relative advantage for deep learners of training with
other classes than those of interest, by comparing learners trained with
62 classes with learners trained with only a subset (on which they
are then tested).
The conclusion discusses
the more general question of why deep learners may benefit so much from 
the self-taught learning framework. Since out-of-distribution data
(perturbed or from other related classes) is very common, this conclusion
is of practical importance.

\section{Perturbed and Transformed Character Images}
\label{s:perturbations}

\begin{wrapfigure}[8]{l}{0.15\textwidth}
\begin{center}
\includegraphics[scale=.4]{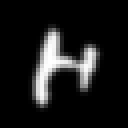}\\
{\bf Original}
\end{center}
\end{wrapfigure}
This section describes the different transformations we used to stochastically
transform $32 \times 32$ source images (such as the one on the left)
in order to obtain data from a larger distribution which
covers a domain substantially larger than the clean characters distribution from
which we start. 
Although character transformations have been used before to
improve character recognizers, this effort is on a large scale both
in number of classes and in the complexity of the transformations, hence
in the complexity of the learning task.
The code for these transformations (mostly python) is available at 
{\tt http://hg.assembla.com/ift6266}. All the modules in the pipeline share
a global control parameter ($0 \le complexity \le 1$) that allows one to modulate the
amount of deformation or noise introduced. 
There are two main parts in the pipeline. The first one,
from slant to pinch below, performs transformations. The second
part, from blur to contrast, adds different kinds of noise.

\subsection{Transformations}

\subsubsection*{Thickness}

\begin{minipage}[b]{0.14\linewidth}
\begin{center}
\vspace*{-5mm}
\includegraphics[scale=.4]{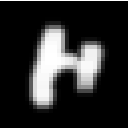}\\
\end{center}
\vspace{.6cm}
\end{minipage}%
\hspace{0.3cm}\begin{minipage}[b]{0.86\linewidth}
To change character {\bf thickness}, morphological operators of dilation and erosion~\citep{Haralick87,Serra82}
are applied. The neighborhood of each pixel is multiplied
element-wise with a {\em structuring element} matrix.
The pixel value is replaced by the maximum or the minimum of the resulting
matrix, respectively for dilation or erosion. Ten different structural elements with 
increasing dimensions (largest is $5\times5$) were used.  For each image, 
randomly sample the operator type (dilation or erosion) with equal probability and one structural
element from a subset of the $n=round(m \times complexity)$ smallest structuring elements
where $m=10$ for dilation and $m=6$ for erosion (to avoid completely erasing thin characters).  
A neutral element (no transformation) 
is always present in the set.
\end{minipage}

\vspace{2mm}

\subsubsection*{Slant}
\vspace*{2mm}

\begin{minipage}[b]{0.14\linewidth}
\centering
\includegraphics[scale=.4]{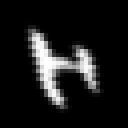}\\
\end{minipage}%
\hspace{0.3cm}
\begin{minipage}[b]{0.83\linewidth}
To produce {\bf slant}, each row of the image is shifted
proportionally to its height: $shift = round(slant \times height)$.  
$slant \sim U[-complexity,complexity]$.
The shift is randomly chosen to be either to the left or to the right.
\vspace{5mm}
\end{minipage}


\subsubsection*{Affine Transformations}

\begin{minipage}[b]{0.14\linewidth}
\begin{center}
\includegraphics[scale=.4]{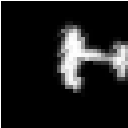}
\vspace*{6mm}
\end{center}
\end{minipage}%
\hspace{0.3cm}\begin{minipage}[b]{0.86\linewidth}
\noindent A $2 \times 3$ {\bf affine transform} matrix (with
parameters $(a,b,c,d,e,f)$) is sampled according to the $complexity$.
Output pixel $(x,y)$ takes the value of input pixel
nearest to $(ax+by+c,dx+ey+f)$,
producing scaling, translation, rotation and shearing.
Marginal distributions of $(a,b,c,d,e,f)$ have been tuned to
forbid large rotations (to avoid confusing classes) but to give good
variability of the transformation: $a$ and $d$ $\sim U[1-3
complexity,1+3\,complexity]$, $b$ and $e$ $\sim U[-3 \,complexity,3\,
complexity]$, and $c$ and $f \sim U[-4 \,complexity, 4 \,
complexity]$.\\
\end{minipage}

\subsubsection*{Local Elastic Deformations}

\begin{minipage}[b]{0.14\linewidth}
\begin{center}
\vspace*{5mm}
\includegraphics[scale=.4]{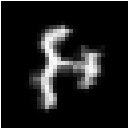}
\end{center}
\end{minipage}%
\hspace{3mm}
\begin{minipage}[b]{0.85\linewidth}
The {\bf local elastic deformation}
module induces a ``wiggly'' effect in the image, following~\citet{SimardSP03-short},
which provides more details. 
The intensity of the displacement fields is given by 
$\alpha = \sqrt[3]{complexity} \times 10.0$, which are 
convolved with a Gaussian 2D kernel (resulting in a blur) of
standard deviation $\sigma = 10 - 7 \times\sqrt[3]{complexity}$.
\vspace{2mm}
\end{minipage}

\vspace*{4mm}

\subsubsection*{Pinch}

\begin{minipage}[b]{0.14\linewidth}
\begin{center}
\includegraphics[scale=.4]{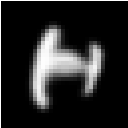}\\
\vspace*{15mm}
\end{center}
\end{minipage}%
\hspace{0.3cm}\begin{minipage}[b]{0.86\linewidth}
The {\bf pinch} module applies the ``Whirl and pinch'' GIMP filter with whirl set to 0. 
A pinch is ``similar to projecting the image onto an elastic
surface and pressing or pulling on the center of the surface'' (GIMP documentation manual).
For a square input image, draw a radius-$r$ disk
around its center $C$. Any pixel $P$ belonging to
that disk has its value replaced by
the value of a ``source'' pixel in the original image,
on the line that goes through $C$ and $P$, but
at some other distance $d_2$. Define $d_1=distance(P,C)$
and $d_2 = sin(\frac{\pi{}d_1}{2r})^{-pinch} \times
d_1$, where $pinch$ is a parameter of the filter.
The actual value is given by bilinear interpolation considering the pixels
around the (non-integer) source position thus found.
Here $pinch \sim U[-complexity, 0.7 \times complexity]$.
\end{minipage}


\subsection{Injecting Noise}

\subsubsection*{Motion Blur}

\begin{minipage}[t]{0.14\linewidth}
\centering
\vspace*{0mm}
\includegraphics[scale=.4]{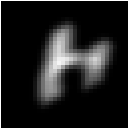}
\end{minipage}%
\hspace{0.3cm}\begin{minipage}[t]{0.83\linewidth}
\vspace*{2mm}
The {\bf motion blur} module is GIMP's ``linear motion blur'', which
has parameters $length$ and $angle$. The value of
a pixel in the final image is approximately the  mean of the first $length$ pixels
found by moving in the $angle$ direction,
$angle \sim U[0,360]$ degrees, and $length \sim {\rm Normal}(0,(3 \times complexity)^2)$.
\end{minipage}


\subsubsection*{Occlusion}

\begin{minipage}[t]{0.14\linewidth}
\centering
\vspace*{3mm}
\includegraphics[scale=.4]{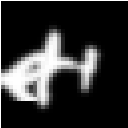}\\
\end{minipage}%
\hspace{0.3cm}\begin{minipage}[t]{0.83\linewidth}
The {\bf occlusion} module selects a random rectangle from an {\em occluder} character
image and places it over the original {\em occluded}
image. Pixels are combined by taking the max(occluder, occluded),
i.e. keeping the lighter ones.
The rectangle corners
are sampled so that larger complexity gives larger rectangles.
The destination position in the occluded image are also sampled
according to a normal distribution.
This module is skipped with probability 60\%.
\end{minipage}

\subsubsection*{Gaussian Smoothing}

\begin{minipage}[t]{0.14\linewidth}
\begin{center}
\vspace*{6mm}
\includegraphics[scale=.4]{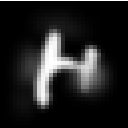}
\end{center}
\end{minipage}%
\hspace{0.3cm}\begin{minipage}[t]{0.86\linewidth}
With the {\bf Gaussian smoothing} module, 
different regions of the image are spatially smoothed.
This is achieved  by first convolving
the image with an isotropic Gaussian kernel of
size and variance chosen uniformly in the ranges $[12,12 + 20 \times
complexity]$ and $[2,2 + 6 \times complexity]$. This filtered image is normalized
between $0$ and $1$.  We also create an isotropic weighted averaging window, of the
kernel size, with maximum value at the center.  For each image we sample
uniformly from $3$ to $3 + 10 \times complexity$ pixels that will be
averaging centers between the original image and the filtered one.  We
initialize to zero a mask matrix of the image size. For each selected pixel
we add to the mask the averaging window centered on it.  The final image is
computed from the following element-wise operation: $\frac{image + filtered\_image
\times mask}{mask+1}$.
This module is skipped with probability 75\%.
\end{minipage}


\subsubsection*{Permute Pixels}

\begin{minipage}[t]{0.14\textwidth}
\begin{center}
\vspace*{1mm}
\includegraphics[scale=.4]{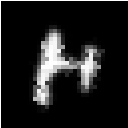}
\end{center}
\end{minipage}%
\hspace{3mm}\begin{minipage}[t]{0.86\linewidth}
\vspace*{1mm}
This module {\bf permutes neighbouring pixels}. It first selects a
fraction $\frac{complexity}{3}$ of pixels randomly in the image. Each
of these pixels is then sequentially exchanged with a random pixel
among its four nearest neighbors (on its left, right, top or bottom).
This module is skipped with probability 80\%.\\
\end{minipage}


\subsubsection*{Gaussian Noise}

\begin{minipage}[t]{0.14\textwidth}
\begin{center}
\vspace*{0mm}
\includegraphics[scale=.4]{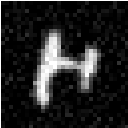}
\end{center}
\end{minipage}%
\hspace{0.3cm}\begin{minipage}[t]{0.86\linewidth}
\vspace*{1mm}
The {\bf Gaussian noise} module simply adds, to each pixel of the image independently, a
noise $\sim Normal(0,(\frac{complexity}{10})^2)$.
This module is skipped with probability 70\%.
\end{minipage}


\subsubsection*{Background Image Addition}

\begin{minipage}[t]{\linewidth}
\begin{minipage}[t]{0.14\linewidth}
\centering
\vspace*{0mm}
\includegraphics[scale=.4]{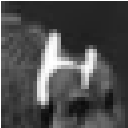}
\end{minipage}%
\hspace{0.3cm}\begin{minipage}[t]{0.83\linewidth}
\vspace*{1mm}
Following~\citet{Larochelle-jmlr-2009}, the {\bf background image} module adds a random
background image behind the letter, from a randomly chosen natural image,
with contrast adjustments depending on $complexity$, to preserve
more or less of the original character image.
\end{minipage}
\end{minipage}

\subsubsection*{Salt and Pepper Noise}

\begin{minipage}[t]{0.14\linewidth}
\centering
\vspace*{0mm}
\includegraphics[scale=.4]{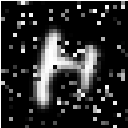}
\end{minipage}%
\hspace{0.3cm}\begin{minipage}[t]{0.83\linewidth}
\vspace*{1mm}
The {\bf salt and pepper noise} module adds noise $\sim U[0,1]$ to random subsets of pixels.
The number of selected pixels is $0.2 \times complexity$.
This module is skipped with probability 75\%.
\end{minipage}

\subsubsection*{Scratches}

\begin{minipage}[t]{0.14\textwidth}
\begin{center}
\vspace*{4mm}
\includegraphics[scale=.4]{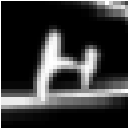}\\
\end{center}
\end{minipage}%
\hspace{0.3cm}\begin{minipage}[t]{0.86\linewidth}
The {\bf scratches} module places line-like white patches on the image.  The
lines are heavily transformed images of the digit ``1'' (one), chosen
at random among 500 such 1 images,
randomly cropped and rotated by an angle $\sim Normal(0,(100 \times
complexity)^2$ (in degrees), using bi-cubic interpolation.
Two passes of a grey-scale morphological erosion filter
are applied, reducing the width of the line
by an amount controlled by $complexity$.
This module is skipped with probability 85\%. The probabilities
of applying 1, 2, or 3 patches are (50\%,30\%,20\%).
\end{minipage}


\subsubsection*{Grey Level and Contrast Changes}

\begin{minipage}[t]{0.15\linewidth}
\centering
\vspace*{0mm}
\includegraphics[scale=.4]{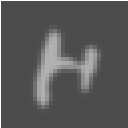}
\end{minipage}%
\hspace{3mm}\begin{minipage}[t]{0.85\linewidth}
\vspace*{1mm}
The {\bf grey level and contrast} module changes the contrast by changing grey levels, and may invert the image polarity (white
to black and black to white). The contrast is $C \sim U[1-0.85 \times complexity,1]$ 
so the image is normalized into $[\frac{1-C}{2},1-\frac{1-C}{2}]$. The
polarity is inverted with probability 50\%.
\end{minipage}

\section{Experimental Setup}

Much previous work on deep learning had been performed on
the MNIST digits task~\citep{Hinton06,ranzato-07-small,Bengio-nips-2006,Salakhutdinov+Hinton-2009},
with 60~000 examples, and variants involving 10~000
examples~\citep{Larochelle-jmlr-toappear-2008,VincentPLarochelleH2008}.
The focus here is on much larger training sets, from 10 times to 
to 1000 times larger, and 62 classes.

The first step in constructing the larger datasets (called NISTP and P07) is to sample from
a {\em data source}: {\bf NIST} (NIST database 19), {\bf Fonts}, {\bf Captchas},
and {\bf OCR data} (scanned machine printed characters). Once a character
is sampled from one of these sources (chosen randomly), the second step is to
apply a pipeline of transformations and/or noise processes described in section \ref{s:perturbations}.

To provide a baseline of error rate comparison we also estimate human performance
on both the 62-class task and the 10-class digits task.
We compare the best Multi-Layer Perceptrons (MLP) against
the best Stacked Denoising Auto-encoders (SDA), when
both models' hyper-parameters are selected to minimize the validation set error.
We also provide a comparison against a precise estimate
of human performance obtained via Amazon's Mechanical Turk (AMT)
service (http://mturk.com). 
AMT users are paid small amounts
of money to perform tasks for which human intelligence is required.
Mechanical Turk has been used extensively in natural language processing and vision.
AMT users were presented
with 10 character images (from a test set) and asked to choose 10 corresponding ASCII
characters. They were forced to choose a single character class (either among the
62 or 10 character classes) for each image.
80 subjects classified 2500 images per (dataset,task) pair.
Different humans labelers sometimes provided a different label for the same
example, and we were able to estimate the error variance due to this effect
because each image was classified by 3 different persons. 
The average error of humans on the 62-class task NIST test set
is 18.2\%, with a standard error of 0.1\%.

\subsection{Data Sources}

{\bf NIST.}
Our main source of characters is the NIST Special Database 19~\citep{Grother-1995}, 
widely used for training and testing character
recognition systems~\citep{Granger+al-2007,Cortes+al-2000,Oliveira+al-2002-short,Milgram+al-2005}. 
The dataset is composed of 814255 digits and characters (upper and lower cases), with hand checked classifications,
extracted from handwritten sample forms of 3600 writers. The characters are labelled by one of the 62 classes 
corresponding to ``0''-``9'',``A''-``Z'' and ``a''-``z''. The dataset contains 8 parts (partitions) of varying complexity. 
The fourth partition (called $hsf_4$, 82587 examples), 
experimentally recognized to be the most difficult one, is the one recommended 
by NIST as a testing set and is used in our work as well as some previous work~\citep{Granger+al-2007,Cortes+al-2000,Oliveira+al-2002-short,Milgram+al-2005}
for that purpose. We randomly split the remainder (731668 examples) into a training set and a validation set for
model selection. 
The performances reported by previous work on that dataset mostly use only the digits.
Here we use all the classes both in the training and testing phase. This is especially
useful to estimate the effect of a multi-task setting.
The distribution of the classes in the NIST training and test sets differs
substantially, with relatively many more digits in the test set, and a more uniform distribution
of letters in the test set (whereas in the training set they are distributed
more like in natural text).

{\bf Fonts.} 
In order to have a good variety of sources we downloaded an important number of free fonts from:
{\tt http://cg.scs.carleton.ca/\textasciitilde luc/freefonts.html}.
Including the operating system's (Windows 7) fonts, there is a total of $9817$ different fonts that we can choose uniformly from.
The chosen {\tt ttf} file is either used as input of the Captcha generator (see next item) or, by producing a corresponding image, 
directly as input to our models.

{\bf Captchas.}
The Captcha data source is an adaptation of the \emph{pycaptcha} library (a python based captcha generator library) for 
generating characters of the same format as the NIST dataset. This software is based on
a random character class generator and various kinds of transformations similar to those described in the previous sections. 
In order to increase the variability of the data generated, many different fonts are used for generating the characters. 
Transformations (slant, distortions, rotation, translation) are applied to each randomly generated character with a complexity
depending on the value of the complexity parameter provided by the user of the data source. 

{\bf OCR data.}
A large set (2 million) of scanned, OCRed and manually verified machine-printed 
characters where included as an
additional source. This set is part of a larger corpus being collected by the Image Understanding
Pattern Recognition Research group led by Thomas Breuel at University of Kaiserslautern 
({\tt http://www.iupr.com}), and which will be publicly released.

\subsection{Data Sets}

All data sets contain 32$\times$32 grey-level images (values in $[0,1]$) associated with a label
from one of the 62 character classes.

{\bf NIST.} This is the raw NIST special database 19~\citep{Grother-1995}. It has
\{651668 / 80000 / 82587\} \{training / validation / test\} examples.

{\bf P07.} This dataset is obtained by taking raw characters from all four of the above sources
and sending them through the transformation pipeline described in section \ref{s:perturbations}.
For each new example to generate, a data source is selected with probability $10\%$ from the fonts,
$25\%$ from the captchas, $25\%$ from the OCR data and $40\%$ from NIST. We apply all the transformations in the
order given above, and for each of them we sample uniformly a \emph{complexity} in the range $[0,0.7]$.
It has \{81920000 / 80000 / 20000\} \{training / validation / test\} examples.

{\bf NISTP.} This one is equivalent to P07 (complexity parameter of $0.7$ with the same proportions of data sources)
  except that we only apply
  transformations from slant to pinch. Therefore, the character is
  transformed but no additional noise is added to the image, giving images
  closer to the NIST dataset. 
It has \{81920000 / 80000 / 20000\} \{training / validation / test\} examples.

\subsection{Models and their Hyperparameters}

The experiments are performed using MLPs (with a single
hidden layer) and SDAs.
\emph{Hyper-parameters are selected based on the {\bf NISTP} validation set error.}

{\bf Multi-Layer Perceptrons (MLP).}
Whereas previous work had compared deep architectures to both shallow MLPs and
SVMs, we only compared to MLPs here because of the very large datasets used
(making the use of SVMs computationally challenging because of their quadratic
scaling behavior). Preliminary experiments on training SVMs (libSVM) with subsets of the training
set allowing the program to fit in memory yielded substantially worse results
than those obtained with MLPs. For training on nearly a billion examples
(with the perturbed data), the MLPs and SDA are much more convenient than
classifiers based on kernel methods.
The MLP has a single hidden layer with $\tanh$ activation functions, and softmax (normalized
exponentials) on the output layer for estimating $P(class | image)$.
The number of hidden units is taken in $\{300,500,800,1000,1500\}$. 
Training examples are presented in minibatches of size 20. A constant learning
rate was chosen among $\{0.001, 0.01, 0.025, 0.075, 0.1, 0.5\}$.

{\bf Stacked Denoising Auto-Encoders (SDA).}
Various auto-encoder variants and Restricted Boltzmann Machines (RBMs)
can be used to initialize the weights of each layer of a deep MLP (with many hidden 
layers)~\citep{Hinton06,ranzato-07-small,Bengio-nips-2006}, 
apparently setting parameters in the
basin of attraction of supervised gradient descent yielding better 
generalization~\citep{Erhan+al-2010}.  This initial {\em unsupervised
pre-training phase} uses all of the training images but not the training labels.
Each layer is trained in turn to produce a new representation of its input
(starting from the raw pixels).
It is hypothesized that the
advantage brought by this procedure stems from a better prior,
on the one hand taking advantage of the link between the input
distribution $P(x)$ and the conditional distribution of interest
$P(y|x)$ (like in semi-supervised learning), and on the other hand
taking advantage of the expressive power and bias implicit in the
deep architecture (whereby complex concepts are expressed as
compositions of simpler ones through a deep hierarchy).

\begin{figure}[ht]
\centerline{\resizebox{0.8\textwidth}{!}{\includegraphics{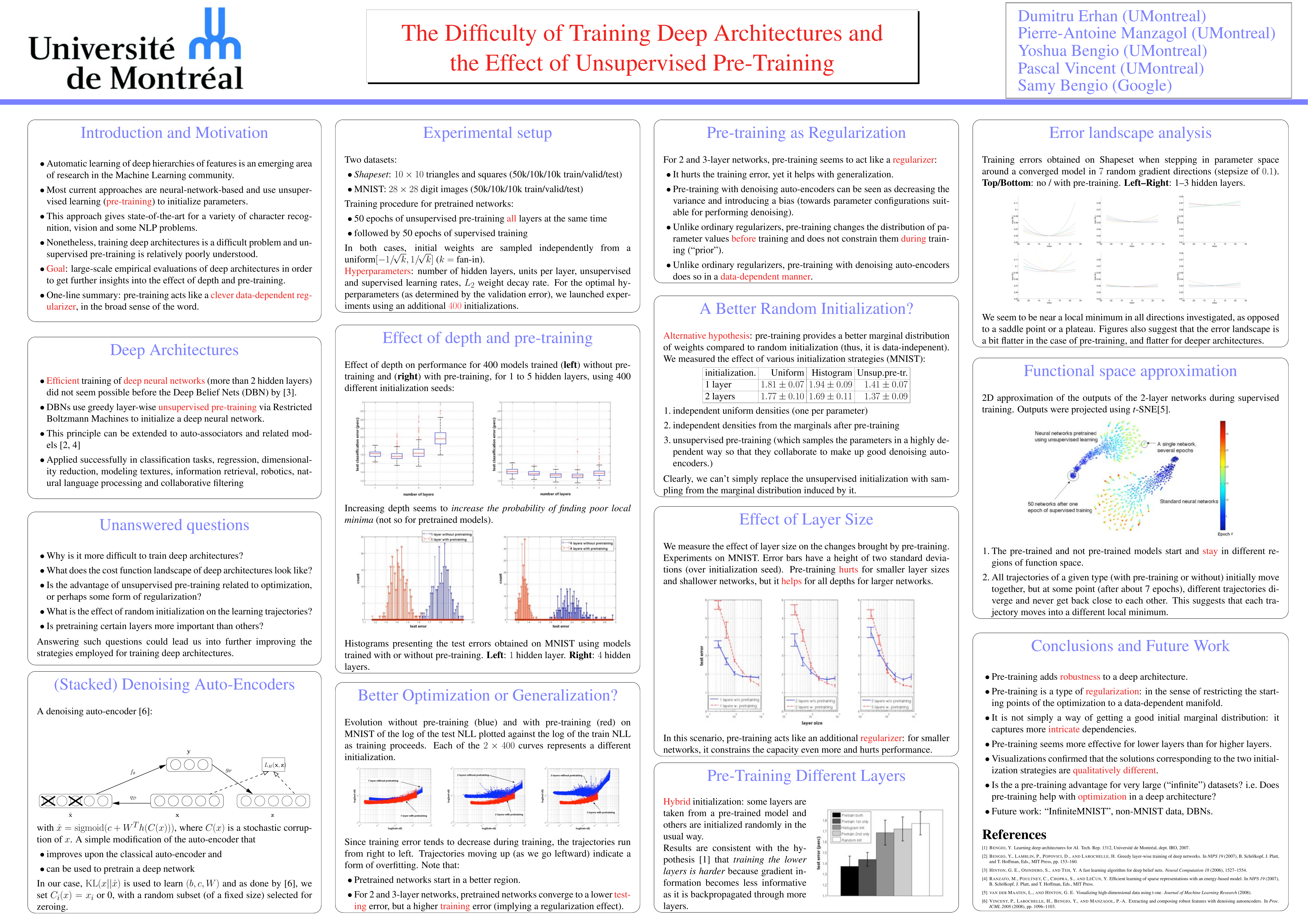}}}
\caption{Illustration of the computations and training criterion for the denoising
auto-encoder used to pre-train each layer of the deep architecture. Input $x$ of
the layer (i.e. raw input or output of previous layer)
s corrupted into $\tilde{x}$ and encoded into code $y$ by the encoder $f_\theta(\cdot)$.
The decoder $g_{\theta'}(\cdot)$ maps $y$ to reconstruction $z$, which
is compared to the uncorrupted input $x$ through the loss function
$L_H(x,z)$, whose expected value is approximately minimized during training
by tuning $\theta$ and $\theta'$.}
\label{fig:da}
\end{figure}

Here we chose to use the Denoising
Auto-encoder~\citep{VincentPLarochelleH2008} as the building block for
these deep hierarchies of features, as it is simple to train and
explain (see Figure~\ref{fig:da}, as well as 
tutorial and code there: {\tt http://deeplearning.net/tutorial}), 
provides efficient inference, and yielded results
comparable or better than RBMs in series of experiments
\citep{VincentPLarochelleH2008}. During training, a Denoising
Auto-encoder is presented with a stochastically corrupted version
of the input and trained to reconstruct the uncorrupted input,
forcing the hidden units to represent the leading regularities in
the data. Here we use the random binary masking corruption
(which sets to 0 a random subset of the inputs).
 Once it is trained, in a purely unsupervised way, 
its hidden units' activations can
be used as inputs for training a second one, etc.
After this unsupervised pre-training stage, the parameters
are used to initialize a deep MLP, which is fine-tuned by
the same standard procedure used to train them (see previous section).
The SDA hyper-parameters are the same as for the MLP, with the addition of the
amount of corruption noise (we used the masking noise process, whereby a
fixed proportion of the input values, randomly selected, are zeroed), and a
separate learning rate for the unsupervised pre-training stage (selected
from the same above set). The fraction of inputs corrupted was selected
among $\{10\%, 20\%, 50\%\}$. Another hyper-parameter is the number
of hidden layers but it was fixed to 3 based on previous work with
SDAs on MNIST~\citep{VincentPLarochelleH2008}. The size of the hidden
layers was kept constant across hidden layers, and the best results
were obtained with the largest values that we could experiment
with given our patience, with 1000 hidden units.


\begin{figure}[ht]
\centerline{\resizebox{.99\textwidth}{!}{\includegraphics{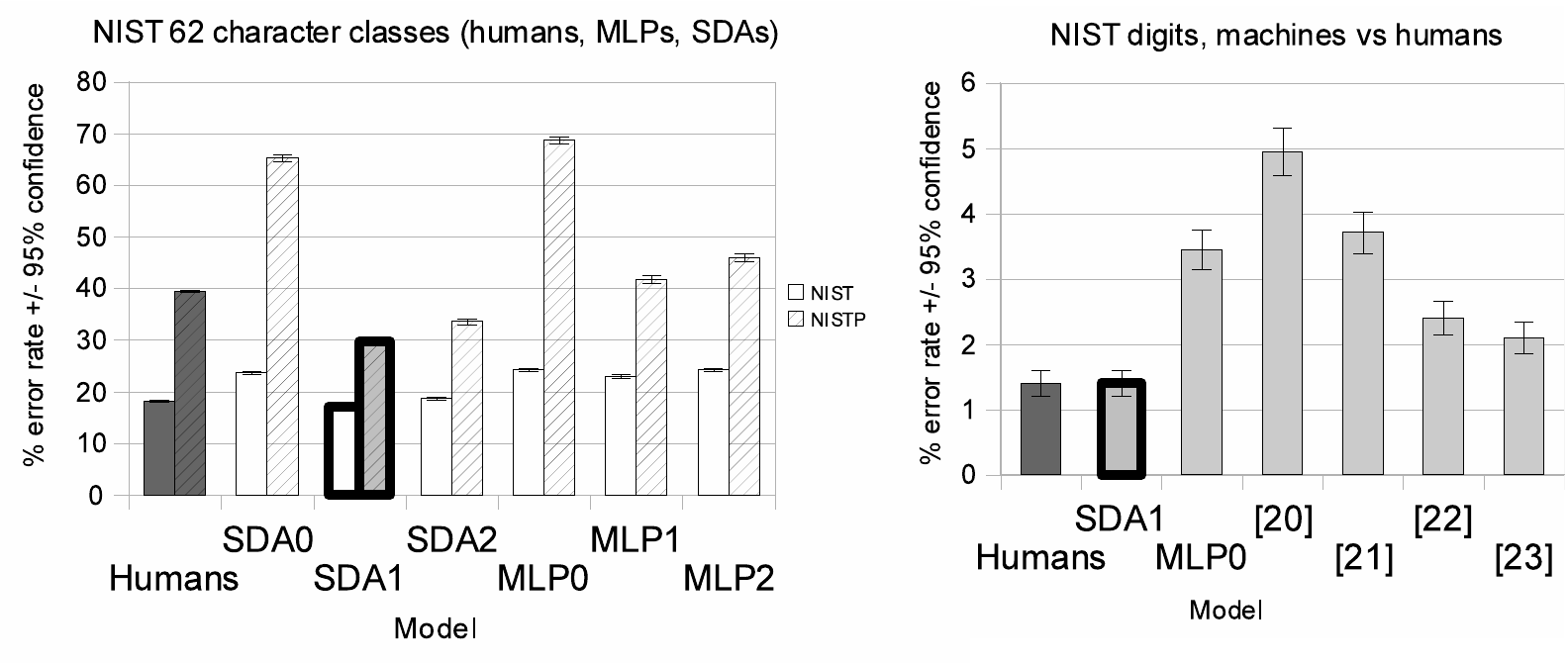}}}
\caption{SDAx are the {\bf deep} models. Error bars indicate a 95\% confidence interval. 0 indicates that the model was trained
on NIST, 1 on NISTP, and 2 on P07. Left: overall results
of all models, on NIST and NISTP test sets.
Right: error rates on NIST test digits only, along with the previous results from 
literature~\citep{Granger+al-2007,Cortes+al-2000,Oliveira+al-2002-short,Milgram+al-2005}
respectively based on ART, nearest neighbors, MLPs, and SVMs.}
\label{fig:error-rates-charts}
\end{figure}

\begin{figure}[ht]
\centerline{\resizebox{.99\textwidth}{!}{\includegraphics{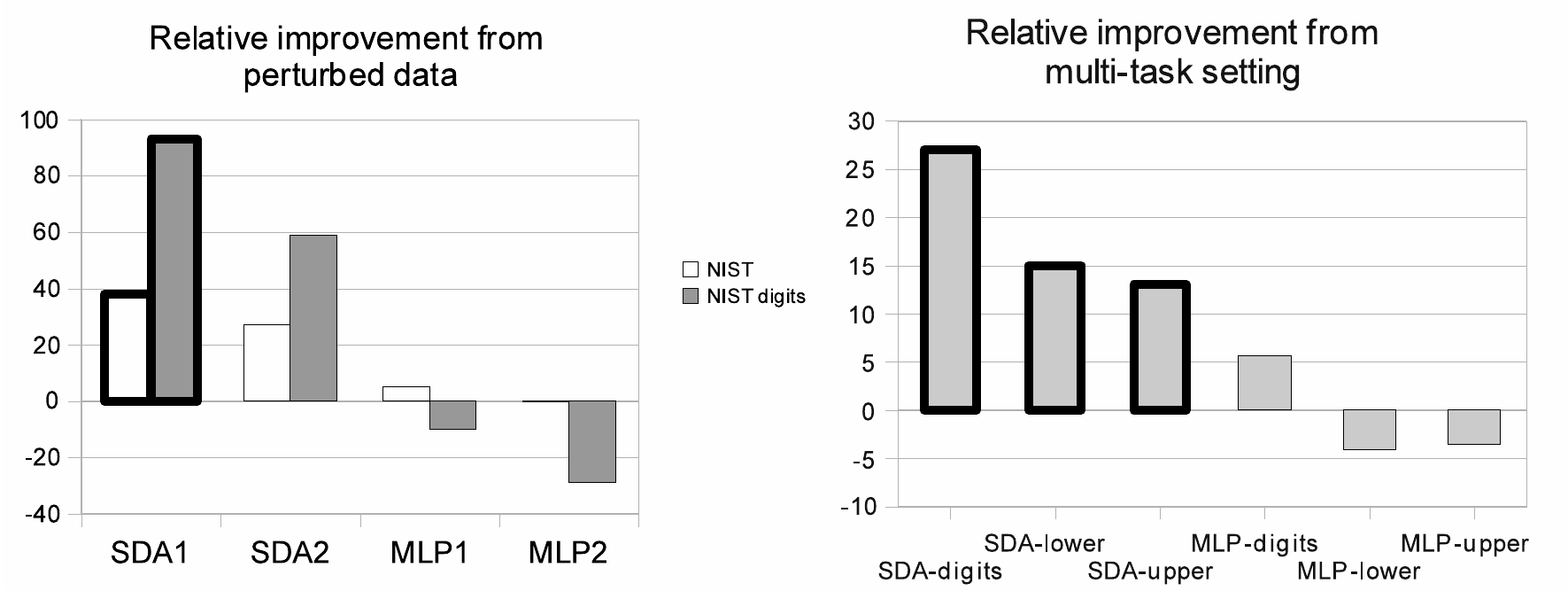}}}
\caption{Relative improvement in error rate due to self-taught learning. 
Left: Improvement (or loss, when negative)
induced by out-of-distribution examples (perturbed data). 
Right: Improvement (or loss, when negative) induced by multi-task 
learning (training on all classes and testing only on either digits,
upper case, or lower-case). The deep learner (SDA) benefits more from
both self-taught learning scenarios, compared to the shallow MLP.}
\label{fig:improvements-charts}
\end{figure}

\section{Experimental Results}

The models are either trained on NIST (MLP0 and SDA0), 
NISTP (MLP1 and SDA1), or P07 (MLP2 and SDA2), and tested
on either NIST, NISTP or P07, either on the 62-class task
or on the 10-digits task. Training (including about half
for unsupervised pre-training, for DAs) on the larger
datasets takes around one day on a GPU-285.
Figure~\ref{fig:error-rates-charts} summarizes the results obtained,
comparing humans, the three MLPs (MLP0, MLP1, MLP2) and the three SDAs (SDA0, SDA1,
SDA2), along with the previous results on the digits NIST special database
19 test set from the literature, respectively based on ARTMAP neural
networks ~\citep{Granger+al-2007}, fast nearest-neighbor search
~\citep{Cortes+al-2000}, MLPs ~\citep{Oliveira+al-2002-short}, and SVMs
~\citep{Milgram+al-2005}.  More detailed and complete numerical results
(figures and tables, including standard errors on the error rates) can be
found in Appendix.
The deep learner not only outperformed the shallow ones and
previously published performance (in a statistically and qualitatively
significant way) but when trained with perturbed data
reaches human performance on both the 62-class task
and the 10-class (digits) task. 
17\% error (SDA1) or 18\% error (humans) may seem large but a large
majority of the errors from humans and from SDA1 are from out-of-context
confusions (e.g. a vertical bar can be a ``1'', an ``l'' or an ``L'', and a
``c'' and a ``C'' are often indistinguishible).

In addition, as shown in the left of
Figure~\ref{fig:improvements-charts}, the relative improvement in error
rate brought by self-taught learning is greater for the SDA, and these
differences with the MLP are statistically and qualitatively
significant. 
The left side of the figure shows the improvement to the clean
NIST test set error brought by the use of out-of-distribution examples
(i.e. the perturbed examples examples from NISTP or P07). 
Relative percent change is measured by taking
$100 \% \times$ (original model's error / perturbed-data model's error - 1).
The right side of
Figure~\ref{fig:improvements-charts} shows the relative improvement
brought by the use of a multi-task setting, in which the same model is
trained for more classes than the target classes of interest (i.e. training
with all 62 classes when the target classes are respectively the digits,
lower-case, or upper-case characters). Again, whereas the gain from the
multi-task setting is marginal or negative for the MLP, it is substantial
for the SDA.  Note that to simplify these multi-task experiments, only the original
NIST dataset is used. For example, the MLP-digits bar shows the relative
percent improvement in MLP error rate on the NIST digits test set 
is $100\% \times$ (single-task
model's error / multi-task model's error - 1).  The single-task model is
trained with only 10 outputs (one per digit), seeing only digit examples,
whereas the multi-task model is trained with 62 outputs, with all 62
character classes as examples.  Hence the hidden units are shared across
all tasks.  For the multi-task model, the digit error rate is measured by
comparing the correct digit class with the output class associated with the
maximum conditional probability among only the digit classes outputs.  The
setting is similar for the other two target classes (lower case characters
and upper case characters).

\section{Conclusions and Discussion}

We have found that the self-taught learning framework is more beneficial
to a deep learner than to a traditional shallow and purely
supervised learner. More precisely, 
the answers are positive for all the questions asked in the introduction.

$\bullet$ 
{\bf Do the good results previously obtained with deep architectures on the
MNIST digits generalize to a much larger and richer (but similar)
dataset, the NIST special database 19, with 62 classes and around 800k examples}?
Yes, the SDA {\em systematically outperformed the MLP and all the previously
published results on this dataset} (the ones that we are aware of), {\em in fact reaching human-level
performance} at around 17\% error on the 62-class task and 1.4\% on the digits,
and beating previously published results on the same data.

$\bullet$ 
{\bf To what extent do self-taught learning scenarios help deep learners,
and do they help them more than shallow supervised ones}?
We found that distorted training examples not only made the resulting
classifier better on similarly perturbed images but also on
the {\em original clean examples}, and more importantly and more novel,
that deep architectures benefit more from such {\em out-of-distribution}
examples. MLPs were helped by perturbed training examples when tested on perturbed input 
images (65\% relative improvement on NISTP) 
but only marginally helped (5\% relative improvement on all classes) 
or even hurt (10\% relative loss on digits)
with respect to clean examples . On the other hand, the deep SDAs
were significantly boosted by these out-of-distribution examples.
Similarly, whereas the improvement due to the multi-task setting was marginal or
negative for the MLP (from +5.6\% to -3.6\% relative change), 
it was quite significant for the SDA (from +13\% to +27\% relative change),
which may be explained by the arguments below.

In the original self-taught learning framework~\citep{RainaR2007}, the
out-of-sample examples were used as a source of unsupervised data, and
experiments showed its positive effects in a \emph{limited labeled data}
scenario. However, many of the results by \citet{RainaR2007} (who used a
shallow, sparse coding approach) suggest that the {\em relative gain of self-taught
learning vs ordinary supervised learning} diminishes as the number of labeled examples increases.
We note instead that, for deep
architectures, our experiments show that such a positive effect is accomplished
even in a scenario with a \emph{large number of labeled examples},
i.e., here, the relative gain of self-taught learning is probably preserved
in the asymptotic regime.

{\bf Why would deep learners benefit more from the self-taught learning framework}?
The key idea is that the lower layers of the predictor compute a hierarchy
of features that can be shared across tasks or across variants of the
input distribution. A theoretical analysis of generalization improvements
due to sharing of intermediate features across tasks already points
towards that explanation~\cite{baxter95a}.
Intermediate features that can be used in different
contexts can be estimated in a way that allows to share statistical 
strength. Features extracted through many levels are more likely to
be more abstract (as the experiments in~\citet{Goodfellow2009} suggest),
increasing the likelihood that they would be useful for a larger array
of tasks and input conditions.
Therefore, we hypothesize that both depth and unsupervised
pre-training play a part in explaining the advantages observed here, and future
experiments could attempt at teasing apart these factors.
And why would deep learners benefit from the self-taught learning
scenarios even when the number of labeled examples is very large?
We hypothesize that this is related to the hypotheses studied
in~\citet{Erhan+al-2010}. Whereas in~\citet{Erhan+al-2010}
it was found that online learning on a huge dataset did not make the
advantage of the deep learning bias vanish, a similar phenomenon
may be happening here. We hypothesize that unsupervised pre-training
of a deep hierarchy with self-taught learning initializes the
model in the basin of attraction of supervised gradient descent
that corresponds to better generalization. Furthermore, such good
basins of attraction are not discovered by pure supervised learning
(with or without self-taught settings), and more labeled examples
does not allow the model to go from the poorer basins of attraction discovered
by the purely supervised shallow models to the kind of better basins associated
with deep learning and self-taught learning.

A Flash demo of the recognizer (where both the MLP and the SDA can be compared) 
can be executed on-line at {\tt http://deep.host22.com}.

\section*{Appendix I: Detailed Numerical Results}

These tables correspond to Figures 2 and 3 and contain the raw error rates for each model and dataset considered.
They also contain additional data such as test errors on P07 and standard errors.

\begin{table}[ht]
\caption{Overall comparison of error rates ($\pm$ std.err.) on 62 character classes (10 digits +
26 lower + 26 upper), except for last columns -- digits only, between deep architecture with pre-training
(SDA=Stacked Denoising Autoencoder) and ordinary shallow architecture 
(MLP=Multi-Layer Perceptron). The models shown are all trained using perturbed data (NISTP or P07)
and using a validation set to select hyper-parameters and other training choices. 
\{SDA,MLP\}0 are trained on NIST,
\{SDA,MLP\}1 are trained on NISTP, and \{SDA,MLP\}2 are trained on P07.
The human error rate on digits is a lower bound because it does not count digits that were
recognized as letters. For comparison, the results found in the literature
on NIST digits classification using the same test set are included.}
\label{tab:sda-vs-mlp-vs-humans}
\begin{center}
\begin{tabular}{|l|r|r|r|r|} \hline
      & NIST test          & NISTP test       & P07 test       & NIST test digits   \\ \hline
Humans&   18.2\% $\pm$.1\%   &  39.4\%$\pm$.1\%   &  46.9\%$\pm$.1\%  &  $1.4\%$ \\ \hline 
SDA0   &  23.7\% $\pm$.14\%  &  65.2\%$\pm$.34\%  & 97.45\%$\pm$.06\%  & 2.7\% $\pm$.14\%\\ \hline 
SDA1   &  17.1\% $\pm$.13\%  &  29.7\%$\pm$.3\%  & 29.7\%$\pm$.3\%  & 1.4\% $\pm$.1\%\\ \hline 
SDA2   &  18.7\% $\pm$.13\%  &  33.6\%$\pm$.3\%  & 39.9\%$\pm$.17\%  & 1.7\% $\pm$.1\%\\ \hline 
MLP0   &  24.2\% $\pm$.15\%  & 68.8\%$\pm$.33\%  & 78.70\%$\pm$.14\%  & 3.45\% $\pm$.15\% \\ \hline 
MLP1   &  23.0\% $\pm$.15\%  &  41.8\%$\pm$.35\%  & 90.4\%$\pm$.1\%  & 3.85\% $\pm$.16\% \\ \hline 
MLP2   &  24.3\% $\pm$.15\%  &  46.0\%$\pm$.35\%  & 54.7\%$\pm$.17\%  & 4.85\% $\pm$.18\% \\ \hline 
\citep{Granger+al-2007} &     &                    &                   & 4.95\% $\pm$.18\% \\ \hline
\citep{Cortes+al-2000} &      &                    &                   & 3.71\% $\pm$.16\% \\ \hline
\citep{Oliveira+al-2002} &    &                    &                   & 2.4\% $\pm$.13\% \\ \hline
\citep{Milgram+al-2005} &      &                    &                   & 2.1\% $\pm$.12\% \\ \hline
\end{tabular}
\end{center}
\end{table}

\begin{table}[ht]
\caption{Relative change in error rates due to the use of perturbed training data,
either using NISTP, for the MLP1/SDA1 models, or using P07, for the MLP2/SDA2 models.
A positive value indicates that training on the perturbed data helped for the
given test set (the first 3 columns on the 62-class tasks and the last one is
on the clean 10-class digits). Clearly, the deep learning models did benefit more
from perturbed training data, even when testing on clean data, whereas the MLP
trained on perturbed data performed worse on the clean digits and about the same
on the clean characters. }
\label{tab:perturbation-effect}
\begin{center}
\begin{tabular}{|l|r|r|r|r|} \hline
      & NIST test          & NISTP test      & P07 test       & NIST test digits   \\ \hline
SDA0/SDA1-1   &  38\%      &  84\%           & 228\%          &  93\% \\ \hline 
SDA0/SDA2-1   &  27\%      &  94\%           & 144\%          &  59\% \\ \hline 
MLP0/MLP1-1   &  5.2\%     &  65\%           & -13\%          & -10\%  \\ \hline 
MLP0/MLP2-1   &  -0.4\%    &  49\%           & 44\%           & -29\% \\ \hline 
\end{tabular}
\end{center}
\end{table}

\begin{table}[ht]
\caption{Test error rates and relative change in error rates due to the use of
a multi-task setting, i.e., training on each task in isolation vs training
for all three tasks together, for MLPs vs SDAs. The SDA benefits much
more from the multi-task setting. All experiments on only on the
unperturbed NIST data, using validation error for model selection.
Relative improvement is 1 - single-task error / multi-task error.}
\label{tab:multi-task}
\begin{center}
\begin{tabular}{|l|r|r|r|} \hline
             & single-task  & multi-task  & relative \\ 
             & setting      & setting     & improvement \\ \hline
MLP-digits   &  3.77\%      &  3.99\%     & 5.6\%   \\ \hline 
MLP-lower   &  17.4\%      &  16.8\%     &  -4.1\%    \\ \hline 
MLP-upper   &  7.84\%     &  7.54\%      & -3.6\%    \\ \hline 
SDA-digits   &  2.6\%      &  3.56\%     & 27\%    \\ \hline 
SDA-lower   &  12.3\%      &  14.4\%    & 15\%    \\ \hline 
SDA-upper   &  5.93\%     &  6.78\%      & 13\%    \\ \hline 
\end{tabular}
\end{center}
\end{table}

\clearpage
{
\bibliography{strings,strings-short,strings-shorter,ift6266_ml,specials,aigaion-shorter}
\bibliographystyle{unsrtnat}
}

\end{document}